\documentclass[10pt,twocolumn,letterpaper]{article}

\usepackage{kotex}

\usepackage{multirow}
\usepackage{xfrac}
\usepackage{cuted}
\usepackage{tabularx}
\usepackage{tikz}

\usepackage{amsmath, amssymb}
\usepackage{algorithm}
\usepackage[noend]{algpseudocode} 
\usepackage{pgfplots} 
\usepackage{verbatim}
\usepackage{mdframed}

\usepackage[pagenumbers]{wacv} 

\definecolor{wacvblue}{rgb}{0.21,0.49,0.74}
\usepackage[pagebackref,breaklinks,colorlinks,allcolors=wacvblue]{hyperref}

\def\wacvPaperID{3131} 
\def\confName{WACV}
\def\confYear{2026}

\title{MUSE: Model-based Uncertainty-aware Similarity Estimation for zero-shot 2D Object Detection and Segmentation}

\author{Sungmin Cho\qquad Sungbum Park\qquad Insoo Oh \\
Netmarble\\
{\tt\small cho\_sm@netamrble.com, spark0916@netamrble.com, ioh@netamrble.com}
}

\begin{document}

\maketitle

\begin{abstract}
In this work, we present \textbf{MUSE} (\textbf{M}odel-based \textbf{U}ncertainty-aware \textbf{S}imilarity \textbf{E}stimation), a training-free framework for model-based zero-shot 2D object detection and segmentation.
First, MUSE incorporates 2D multi-view templates from 3D unseen objects and 2D object proposals from the input query image, respectively.
In the embedding stage, we propose a new feature embedding scheme which integrates class and patch embeddings.
Specifically, the patch embeddings are normalized using the generalized mean pooling (GeM).
In the matching stage, a joint similarity score is introduced, which integrates an absolute score and a relative score.
Finally, we update the similarity score using an uncertainty-aware object prior. 
MUSE achieves state-of-the-art performance on the BOP Challenge 2025, ranking first in the Classic Core, H3, and Industrial tracks—without any additional training or fine-tuning.
Therefore, we believe that MUSE is a promising framework for zero-shot 2D object detection and segmentation.
\end{abstract}

\begin{figure}[t]
    \begin{minipage}[b]{1.\linewidth}
        \centering
        \includegraphics[width=1.0\linewidth]{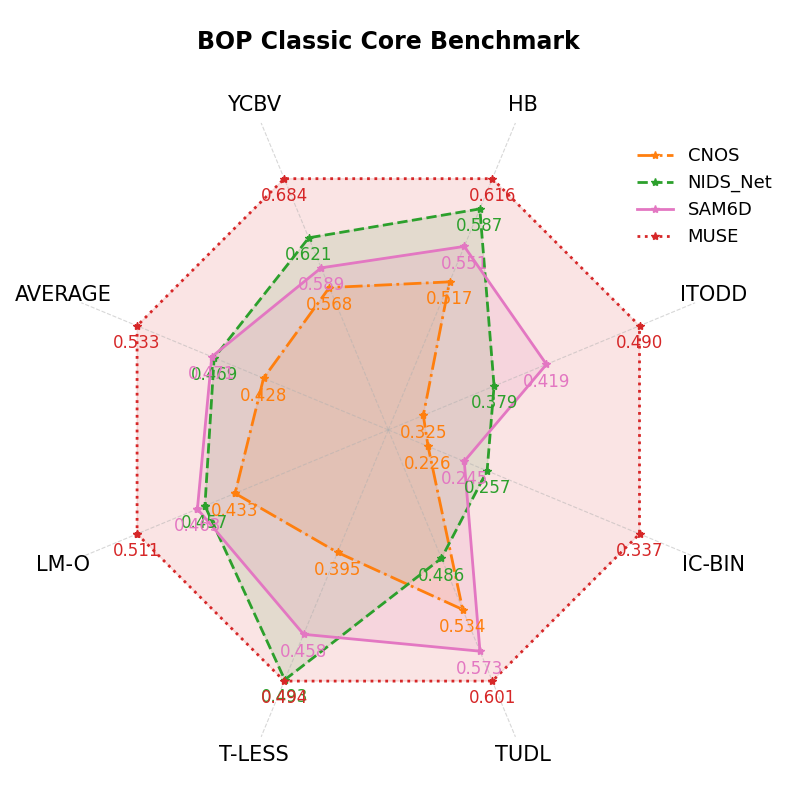}
    \end{minipage}
        \vspace{-3mm}
    %
    \caption{Comparison of detection performance on the BOP Classic Core benchmark.
    Each axis corresponds to a dataset, and values are mAP scores.
    It is demonstrated that MUSE (dotted red) consistently outperforms other methods in all datasets.}
    \label{fig:quantitative}
    \vspace{-3mm}
\end{figure}

\section{Introduction}
\label{sec:intro}

Recent advances in vision-language models~\cite{liu2023grounding, kirillov2023segment, ravi2024sam} and large-scale pretraining models~\cite{caron2021emerging, oquab2023dinov2, simeoni2025dinov3} have enabled impressive generalization to unseen categories.
%
%
However, recognizing and localizing object instances is still challenging in real-world applications such as robotics~\cite{kappler2018real, liu2023robotic, wen2022catgrasp}, augmented reality~\cite{marchand2015pose, manawadu2024advancing}, and industrial inspection~\cite{quentin2023industrial, liu2025exploring}.
%
To address this, the BOP Challenge~\cite{hodan2018bop, Sundermeyer_2023_CVPR, hodan2023bop, hodan2024bop} has introduced rigorous benchmarks for model-based zero-shot 2D object detection and segmentation.
Usually, in model-based zero-shot 2D object detection and segmentation, there are four stages: Onboarding, proposal, embedding, and matching, respectively.
The onboarding stage provides 2D templates from 3D model objects through predefined multi-view camera points.
In the proposal stage, candidate proposals are generated from a query image. 
Specifically, Grounding DINO~\cite{liu2023grounding} and SAM~\cite{kirillov2023segment} generate candidate object proposals for matching.
%
In the embedding stage, a vision foundation model such as DINOv2~\cite{oquab2023dinov2} converts the templates into template features. 
Using the same model, the proposals are also converted into query features for feature matching.
In the matching stage, finally, template features and query features are matched using a similarity metric like the cosine similarity metric. 
The resulting similarity scores are aggregated across all camera views using simple heuristics such as top-K averaging, providing final predictions for all object candidates.

%
%
%
%

As 3D object models and query images are not used in training, model-based zero-shot object recognition is very challenging.
Previous methods~\cite{nguyen2023cnos, lin2024sam, lu2024adapting} typically rely solely on cosine similarity between each proposal and all templates to identify the true class in the zero-shot setting. 
However, these approaches compute raw (absolute) similarity scores for each class independently, without considering the relative differences among classes like other end-to-end classification~\cite{simonyan2014very, krizhevsky2012imagenet}. 
When the scores for different classes are close to each other, the prediction often fails to identify the true class. 
Because each proposal belongs to exactly one class, it is therefore reasonable to consider relative differences among classes in order to obtain more accurate results.
%

Also, in the matching stage, the methods in~\cite{nguyen2023cnos, lin2024sam, lu2024adapting} do not utilize to the proposal confidence which is provided by Grounding DINO~\cite{liu2023grounding} and SAM~\cite{kirillov2023segment}.
As the confidence indicates the reliability of the object proposal, it could be an object prior.
Specifically, the proposal confidence is helpful in scenarios with cluttered backgrounds.
Therefore, in this work, we improve the similarity score by weighting it with the proposal confidence.

To overcome challenges in model-based zero-shot object recognition, we propose MUSE (\textbf{M}odel-based \textbf{U}ncertainty-aware \textbf{S}imilarity \textbf{E}stimation), a model-based zero-shot detection framework that leverages pretrained vision foundation models without any additional training or fine-tuning.
First, we introduce a novel feature integration scheme to handle class embeddings and patch embeddings together.
Specifically, patch embeddings are aggregated using the generalized mean pooling (GeM).
Then, they are combined with class embeddings, yielding new feature embeddings.
Second, a novel joint similarity score metric is presented. 
The joint similarity score combines absolute and relative similarities together.
The joint similarity score enables explicit reasoning over competing candidates~\cite{zhang2018weighted}.
Also, it improves the recognition of hard negative cases. 
Finally, we propose an uncertainty-aware similarity score metric by weighting the proposal confidence within a principled Bayesian framework.
Through extensive evaluation, MUSE achieves robust and accurate zero-shot detection and segmentation, even in cluttered and visually ambiguous scenarios. 

In summary, the main contributions of this work are fourfold:  
\begin{itemize} 
    \item We propose a novel feature integration scheme between class embeddings and patch embeddings using the generalized mean pooling (GeM). 
    \item We introduce a joint similarity metric that combines absolute and relative similarities, enabling more effective discrimination even in hard negative candidates.  
    \item We propose an uncertainty-aware objectness prior using a principled Bayesian framework.
    \item We achieve state-of-the-art results on the BOP Challenge 2025 at the time of submission, specifically in the Classic-Core, H3, and Industrial tracks.  
\end{itemize}

\begin{figure*}[t]
    \begin{minipage}[b]{1.\linewidth}
        \centering
        \includegraphics[width=1.0\linewidth]{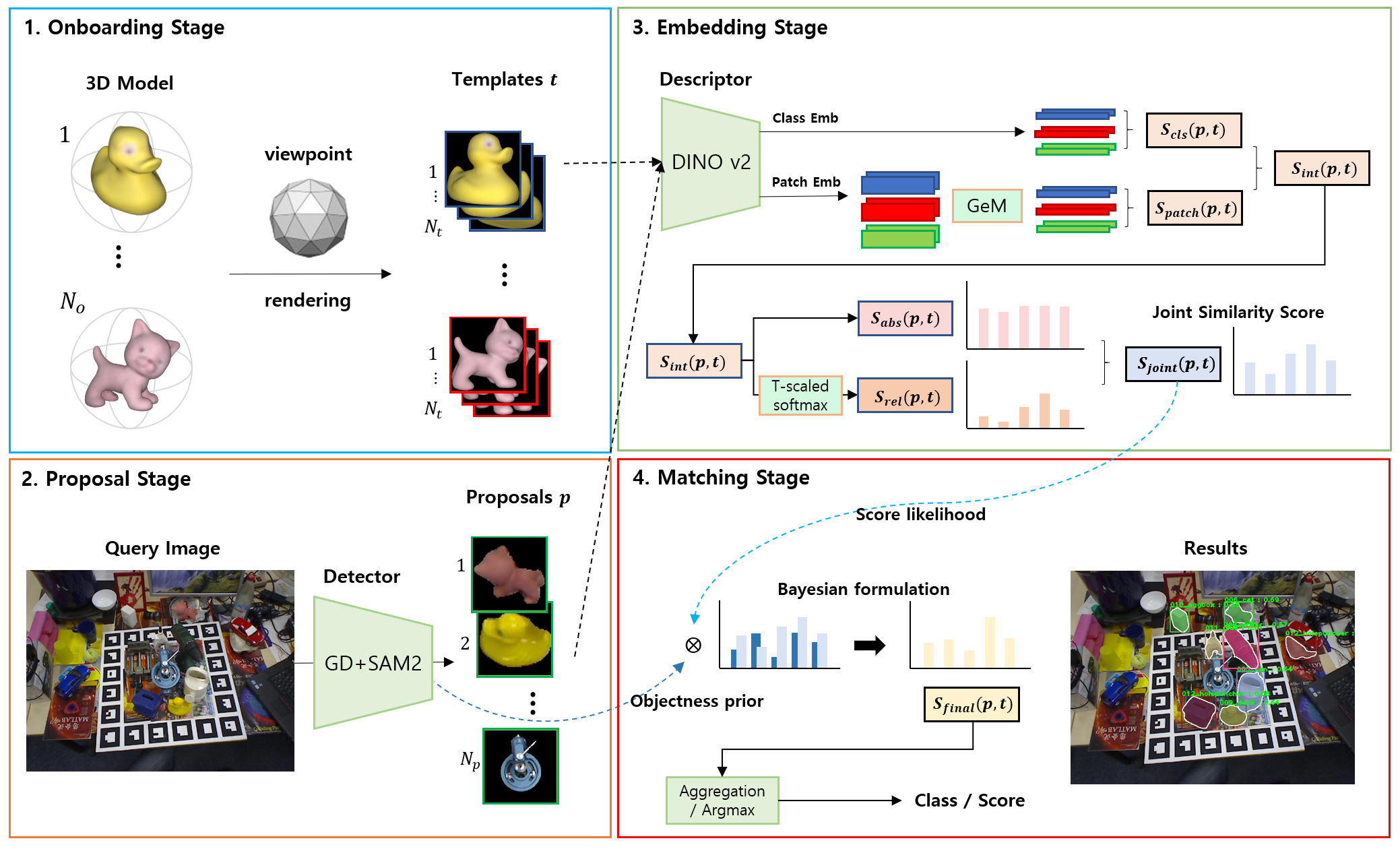}
        \vspace{-6mm}
    \end{minipage}
    \vspace{-3mm}
    \caption{Overview of the proposed framework. Templates and proposals are obtained in the onboarding stage and the proposal stage, respectively. Then, DINOv2~\cite{oquab2023dinov2} yields class and patch embeddings from both templates and proposals. 
    To integrate class and patch embedding into similarity metric, we introduce a generalized mean pooling (GeM). 
    Finally, we further modify the similarity metric using both a joint similarity score scheme and an uncertainty-aware objectness prior, which yields state-of-the-art object detection results.}
    \label{fig:pipeline}
    \vspace{-3mm}
\end{figure*}

\section{Related Work}
\label{sec:related_work}
\subsection{Vision Foundation Model}
\label{subsec2-2}
Vision foundation models have demonstrated strong generalization across diverse vision tasks~\cite{radford2021learning, kirillov2023segment, liu2023grounding, oquab2023dinov2, wang2025vggt, simeoni2025dinov3, bai2023qwen}.
Specifically, Segment Anything (SAM)~\cite{kirillov2023segment}, Segment Anything 2 (SAM 2)~\cite{ravi2024sam}, Grounding DINO~\cite{liu2023grounding}, and DINOv2~\cite{oquab2023dinov2} have been widely used in model-based recognition tasks due to their robust segmentation and representation capabilities.
SAM~\cite{kirillov2023segment} enables a prompt-based segmentation across a wide range of visual concepts.
DINOv2~\cite{oquab2023dinov2} provides strong class-level embeddings without task-specific supervision. 
Grounding DINO~\cite{liu2023grounding} further extends DINO~\cite{liu2023grounding} with open-vocabulary grounding capabilities.

Recent models also have emerged to improve performance or efficiency~\cite{ke2024segment, zhao2023fast, yuan2021florence, xiao2024florence}.
For example, SAM-HQ~\cite{ke2024segment} introduces high-quality masks.
FastSAM~\cite{zhao2023fast} tries to optimize the original network for fast CNN-based segmentation.
Florence models~\cite{yuan2021florence, xiao2024florence} also propose modified multi-modal networks for large-scale vision-language understanding.
As our framework mainly focuses on a similarity metric between feature embeddings, MUSE is easily applicable to vision foundation models.

\subsection{Model-based Zero-Shot Object Detection and Segmentation}
The BOP Challenge~\cite{hodan2023bop} introduces several benchmarks regarding detection and segmentation of unseen objects.
Specifically, the target objects are not shown in the training.
In inference, they predict objects using both 3D models and pretrained vision encoders.
Recently, several model-based zero-shot recognition works ~\cite{nguyen2023cnos, lin2024sam, lu2024adapting} adopt a common pipeline:
They project 3D models into 2D templates through multiple camera viewpoints.
From a query image, then, object proposals are extracted using detectors such as SAM~\cite{kirillov2023segment} or Grounding DINO~\cite{liu2023grounding}.
Then, feature embeddings are extracted from both templates and proposals using pretrained models like DINOv2~\cite{oquab2023dinov2}.
Finally, template embeddings and proposal embeddings are matched using a similarity metric like cosine similarity.

Based on the common pipeline, CNOS~\cite{nguyen2023cnos} compares DINOv2 embeddings of proposals and templates. 
In SAM6D~\cite{lin2024sam}, they combine semantic and geometric cues.
Also, a learnable adapter module is introduced in NIDS-Net~\cite{lu2024adapting}, while they still depend on proposal-template similarity using vision foundation models.
In this work, we introduce a joint similarity score that combines both absolute and relative scores to improve matching robustness even in hard negative cases. 
Then, we integrate the object proposal confidence into the similarity score metric to further enhance matching performance, which is provided in object detectors like SAM~\cite{kirillov2023segment} or Grounding DINO~\cite{liu2023grounding}. 
In summary, sharing the common model-based zero-shot pipeline, we differentiate our approach by modifying the similarity score metric, enabling robust performance for unseen object detection and segmentation.

\section{Proposed Method}
\label{sec:method}
In this section, we present the core components of MUSE framework.
First, class embeddings and patch embeddings are merged together using the generalized mean pooling~\cite{radenovic2018fine}.
Then, we introduce a joint similarity score metric that observes an absolute similarity score and a relative score, improving matching in hard situation like hard negative cases.
Finally, an uncertainty-aware objectness prior leverages proposal confidence within a Bayesian inference framework.

\subsection{Pipeline Overview}
The overall MUSE pipeline is illustrated in Fig.~\ref{fig:pipeline}. 
In the onboarding stage, multiview images are rendered from 3D CAD models as templates. 
Then, a cascaded network of Grounding DINO~\cite{liu2023grounding} and SAM2~\cite{ravi2024sam} predicts object candidate bounding boxes and corresponding candidate regions, respectively, in the proposal stage.
In the embedding stage, DINOv2~\cite{oquab2023dinov2} extracts both class and patch-level embeddings from proposals and templates, respectively.
Note that both embeddings are integrated into a unified object representation using a generalized mean pooling (GeM).
Finally, in the matching stage, proposal embeddings are compared with template embeddings using Tanimoto-based similarity~\cite{tanimoto1957ibm}.
The similarity values are aggregated with the joint similarity score scheme and the uncertainty-aware objectness prior, yielding final predictions.

\begin{figure}[t]
    \centering
    \begin{minipage}[b]{0.11\linewidth}
        \centering
        \scriptsize
        proposals
    \end{minipage}
    \begin{minipage}[b]{0.12\linewidth}
        \centering
        \scriptsize
        cls
    \end{minipage}
    \begin{minipage}[b]{0.11\linewidth}
        \centering
        \scriptsize
        patch
    \end{minipage}
    \begin{minipage}[b]{0.13\linewidth}
        \centering
        \scriptsize
        ours
    \end{minipage}
    \begin{minipage}[b]{0.11\linewidth}
        \centering
        \scriptsize
        proposals
    \end{minipage}
    \begin{minipage}[b]{0.12\linewidth}
        \centering
        \scriptsize
        cls
    \end{minipage}
    \begin{minipage}[b]{0.11\linewidth}
        \centering
        \scriptsize
        patch
    \end{minipage}
    \begin{minipage}[b]{0.12\linewidth}
        \centering
        \scriptsize
        ours
    \end{minipage}
    
    \includegraphics[width=1.0\linewidth]{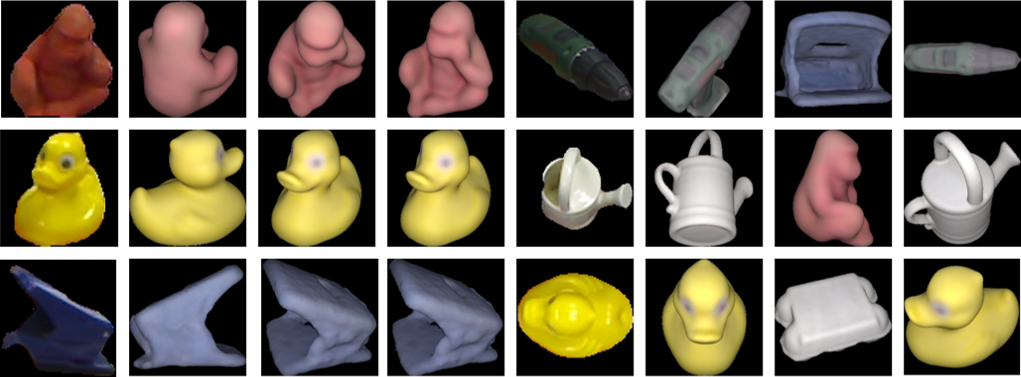}
    \caption{Visualization of the top-1 retrieved templates with respect to embedding types at the LM-O dataset. 
    Each column shows the retrieval templates for class embeddings (cls), patch embeddings (patch), and our integrated embeddings between cls and patch, respectively. 
    It is demonstrated that the proposed integrated embedding method shows both class-level and geometry-level consistencies.}
    \vspace{-4mm}
    \label{fig:top1}
\end{figure}

\subsection{Feature Integration with GeM}
\label{subsec:feature_integration_gem}
In the embedding stage, DINOv2~\cite{oquab2023dinov2} extracts both class embeddings $f^{\mathrm{cls}} \in \mathbb{R}^d$ and patch embeddings $f^{\mathrm{patch}} \in \mathbb{R}^{L\times d}$, where $d$ is the dimension of embedding vector and $L$ is the token length. 
Class embedding provides a global description of an object.
Also, patch embedding keeps fine-grained local details of an object.
In this work, to consider a global and local features together, we propose a novel embedding merging scheme that integrates both class and patch embedding.

As the size of patch embedding is huge and computationally complex, we adopt a generalized mean pooling (GeM)~\cite{radenovic2018fine} for compact patch embedding representation as,
\begin{equation}
\mathrm{GeM}(f^{\mathrm{patch}}) 
= \left( \frac{1}{L} \sum_{l=1}^{L} \left(\{f^{\mathrm{patch}}\}_l\right)^{e} \right)^{\tfrac{1}{e}} \in \mathbb{R}^d,
\end{equation}
where $e$ controls the pooling behavior and $e=1.5$ in our work. and $\{f^{patch}\}_l$ denotes l-th token embeddings of $f^{\mathrm{patch}}$.
Then, the class embedding similarity $S_{\mathrm{cls}}(p,t)$ and the patch embedding similarity metric $S_{\mathrm{patch}}(p,t)$ is defined as,
\begin{equation}
S_{\mathrm{cls}}(p,t)
= \frac{\langle f^{\mathrm{cls}}_{p}, f^{\mathrm{cls}}_{t} \rangle}
{\| f^{\mathrm{cls}}_{p} \| \, \| f^{\mathrm{cls}}_{t} \|},
\end{equation}
\begin{equation}
S_{\mathrm{patch}}(p,t)
= \frac{\langle \mathrm{GeM}(f^{\mathrm{patch}}_{p}), \, \mathrm{GeM}(f^{\mathrm{patch}}_{t}) \rangle}
{\| \mathrm{GeM}(f^{\mathrm{patch}}_{p}) \| \, \| \mathrm{GeM}(f^{\mathrm{patch}}_{t}) \|},
\end{equation}
where $p$ denotes a proposal and $t$ is a template, $ f^{\mathrm{*}}_{p}$, $f^{\mathrm{*}}_{t}$ means the embeddings of * with respect to patch $p$ and template $t$ each.
Also, $\langle \cdot, \cdot \rangle$ is the Tanimoto similarity~\cite{tanimoto1957ibm}.
Finally, the integrated similarity $S_{\mathrm{int}}(p, t) $ is expressed as,
\begin{equation}
S_{\mathrm{int}}(p, t) 
= \alpha \, S_{\mathrm{cls}}(p,t) + (1 - \alpha) \, S_{\mathrm{patch}}(p,t),
\label{eq:score_int}
\end{equation}
where \(\alpha \in [0,1]\) is the weight between the class similarity $S_{\mathrm{cls}}(p,t)$ and the patch similarity $S_{\mathrm{patch}}(p,t)$.
While cosine similarity is commonly used in prior works~\cite{nguyen2023cnos, lin2024sam, lu2024adapting}, we adapt Tanimoto similarity to capture both directional and magnitude information~\cite{tanimoto1957ibm}.
Empirically, the Tanimoto similarity leads more discriminative retrieval.

In Fig.~\ref{fig:top1}, the top-1 retrieved templates from proposals are illustrated with respect to class embedding (cls), patch embedding, and our integrated embedding scheme (ours), respectively.
As the proposed feature integration scheme considers both global and local characteristics of templates, it provides more robust predictions than cases using only class embedding or patch embedding.
As shown in Table~\ref{tab:embedding-ablation}, GeM pooling achieves the best balance among integration strategies. 
In datasets with many objects such as HOT3D (33) and HANDAL (40), patch embeddings require over 1.3GB of VRAM, making memory costs quickly prohibitive. 
In contrast, GeM reduces usage to just 5.5MB while attaining the highest $AP_{\text{core}}$ and $AP_{\text{industrial}}$, demonstrating its effectiveness in compressing patch embeddings into compact descriptors for scalable deployment. 

\begin{table}[t]
\centering

\small
\begin{tabular}{l|c|c|c|c}
\toprule
Method & Pool & \(AP_{core}\) & \(AP_{industrial}\)& VRAM \\
\midrule
Cls                           & -    & 0.478  & 0.290 & \textbf{2 KB} \\
Patch                         & -    & 0.442  & 0.290 & 1 MB \\
\hline
Cls + Patch                   & mean & 0.489  & 0.296 & 4 KB \\
Cls + Patch                   & max  & 0.481  & 0.291 & 4 KB \\
\hline
Cls + Patch                   & GeM & \textbf{0.496} & \textbf{0.305} & 4 KB \\
\bottomrule
\end{tabular}
\caption{Ablation study on different embedding strategies and pooling methods. 
Results are reported as mean Average Precision (mAP) of detection task on the BOP Classic Core ($AP_{\text{core}}$) and Industrial ($AP_{\text{industrial}}$) subsets, with GPU memory usage (VRAM) per objects. 
Our integrated GeM pooling yields the best trade-off between accuracy and efficiency.}
\vspace{-4mm}
\label{tab:embedding-ablation}
\end{table}

\subsection{Joint Similarity Score}
\label{subsec:joint_similarity_score}
In this work, we directly define the absolute score $S_{\mathrm{abs}}(p,t)$ the same as the integrated similarity $S_{\mathrm{int}}(p,t)$.
As $S_{\mathrm{abs}}(p,t)$ refers to only one proposal, it is sensitive to hard negative cases, where several proposals are visually similar to the target template.
To address this, we introduce a relative similarity score by normalizing absolute similarities across all classes using a temperature-scaled softmax as,
\begin{equation}
S_{\mathrm{rel}}(p,t,c) \;=\;
\frac{\exp\!\left( \tfrac{S_{\mathrm{abs}}(p,t,c)}{\tau} \right)}
{\sum_{c'=1}^{N_c} \exp\!\left( \tfrac{S_{\mathrm{abs}}(p,t,c')}{\tau} \right)},
\end{equation}
where \(\tau > 0\) controls the sharpness of the distribution and $N_c$ is the number of classes. 
%
For notational simplicity, the class-wise score $S_{\mathrm{rel}}(p,t,c)$ is expressed as $S_{\mathrm{rel}}(p,t)$.
Finally, the joint similarity score $S_{\mathrm{joint}}(p,t)$ is derived as,
\begin{equation}
S_{\mathrm{joint}}(p,t) 
= \beta \, S_{\mathrm{abs}}(p,t) + (1 - \beta) \, S_{\mathrm{rel}}(p,t),
\end{equation}
where $\beta \in [0,1]$ is the weight between absolute score $S_{\mathrm{abs}}(p,t)$ and the relative score $S_{\mathrm{rel}}(p,t)$.

\begin{figure}[t]
  \centering
  \includegraphics[width=0.99\linewidth]{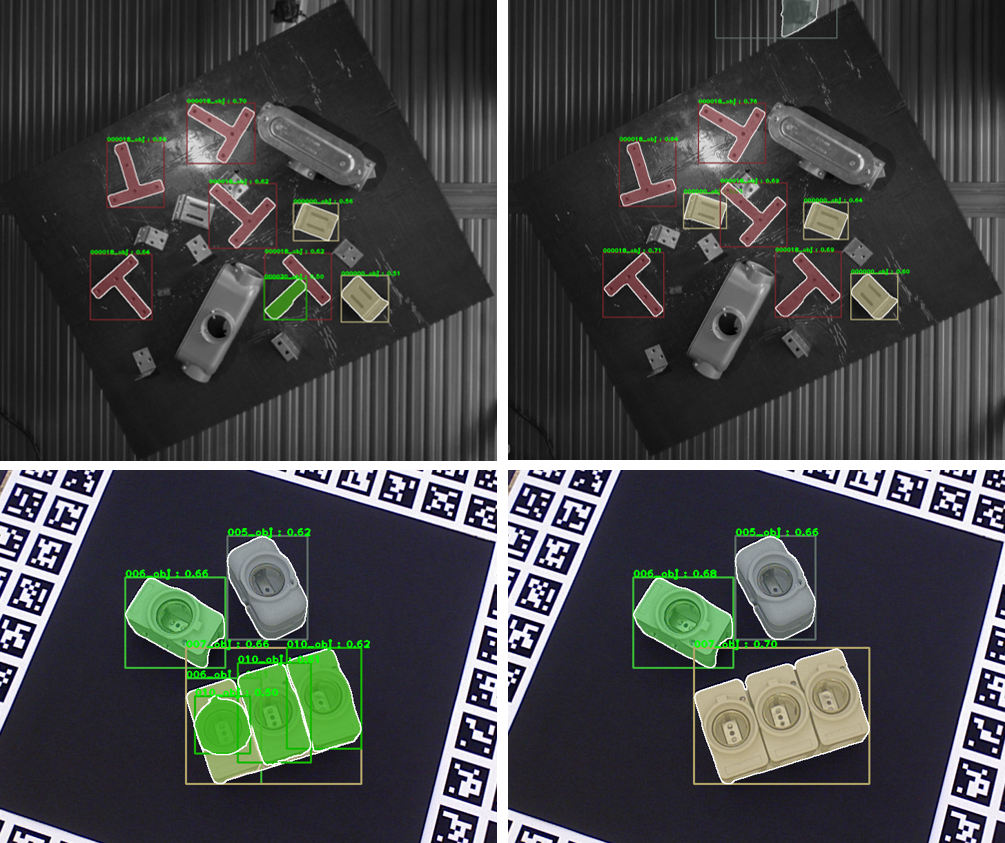}

  \begin{minipage}[b]{0.49\linewidth}
    \centering
    (a) w/o Joint Score
  \end{minipage}
  \begin{minipage}[b]{0.49\linewidth}
    \centering
    (b) w Joint Score
  \end{minipage}

  \caption{Effect of the joint similarity score. The joint similarity score efficiently predicts objects even in the cluttered scene (see top row). Also, visually similar objects (\texttt{obj6} vs. \texttt{obj7}) are clearly separated (see bottom row).}  
  \vspace{-4mm}
  \label{fig:joint-sim}
\end{figure}

In Fig.~\ref{fig:joint-sim}, the effect of the joint similarity score is illustrated.
Compared with the results in Fig.~\ref{fig:joint-sim}~(a), the proposed joint similarity score effectively suppresses hard negative cases, shown in Fig.~\ref{fig:joint-sim}~(b).
Specifically, at the top row in Fig.~\ref{fig:joint-sim}, the mis-detected object (shaded green) is suppressed. 
Additionally, an additional object (shaded yellow) is found even in the cluttered scene.
At the bottom row in Fig.~\ref{fig:joint-sim}, the absolute similarity score fails to discriminate \texttt{obj6} and \texttt{obj7}, which are visually similar.
With the joint similarity score, however, the proposed scheme effectively separates \texttt{obj6} and \texttt{obj7}, shown in the bottom image in Fig.~\ref{fig:joint-sim}~(b).
Our approach, which integrates absolute and relative scores through the joint similarity score, significantly improves discrimination between visually similar objects and suppresses hard negative cases.

\vspace{3mm}

\subsection{Uncertainty-Aware Objectness Prior}
\label{subsec:uncertainty-aware_objectness_prior}
In the proposal stage, Grounding DINO~\cite{liu2023grounding} provides the objectness confidence score $P(O\mid p)$ for each object proposal.
The objectness confidence is used as the prior probability in this work.
Also, we consider the joint similarity score $S_{\mathrm{joint}}(p,t)$ as the class-conditional likelihood $P(C \mid p, O)$. 
By a simple Bayesian formulation~\cite{bayes1958essay}, the posterior is derived as,
\begin{equation}
P(C \mid p) \propto P(O \mid p) \cdot P(C \mid p, O).
\end{equation}
As Grounding DINO~\cite{liu2023grounding} is prompted with generic terms such as \texttt{"items"}, the original objectness score often yields small value. 
To boost up the confidence score, the scaled confidence score $P_{\mathrm{scaled}}(O \mid p)$ is defined as,
\begin{equation}
P_{\mathrm{scaled}}(O \mid p) = \left[P(O \mid p)\right]^{\gamma}.
\end{equation}
where $\gamma \in (0,1)$ is the scale parameter for the prior. 
Therefore, we update the $P(C\mid p)$ as,
\begin{equation}
P(C \mid p) \propto  P_{\mathrm{scaled}}(O \mid p)\cdot S_{\mathrm{joint}}(p,t).
\label{eq:prob}
\end{equation}
Also, the final unnormalized score is derived as,
\begin{equation}
S_{\mathrm{final}}(p,t) = P_{\mathrm{scaled}}(O \mid p) \cdot S_{\mathrm{joint}}(p,t).
\label{eq:final_score}
\end{equation}

\begin{figure}[b]
  \centering
  \includegraphics[width=0.99\linewidth]{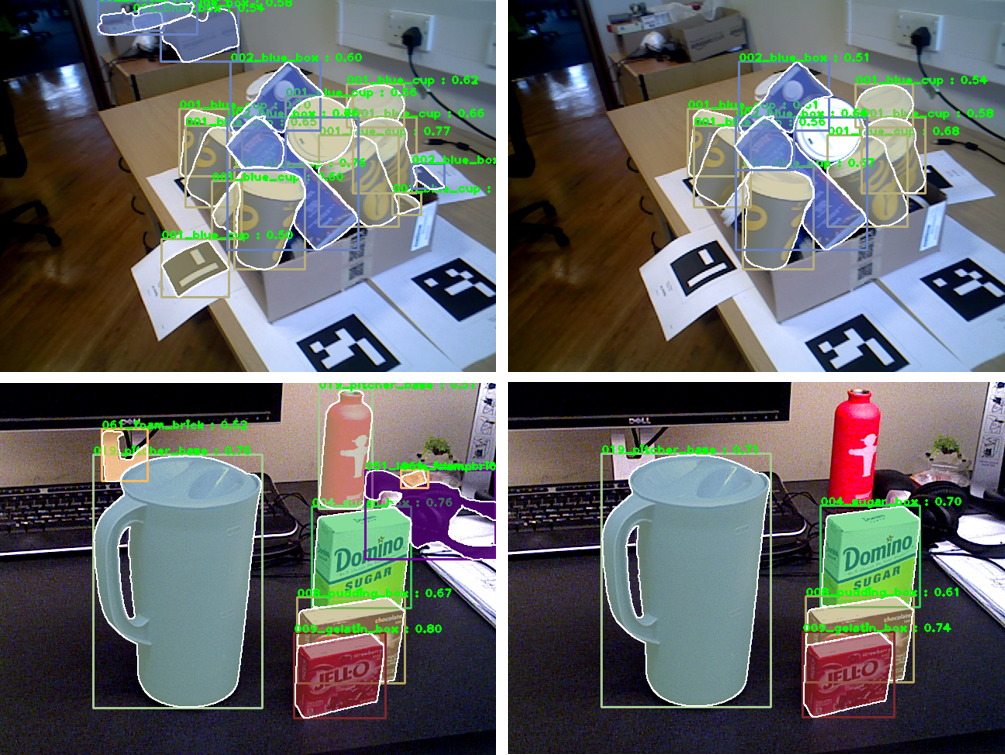}

  \vspace{0.3em}

  \begin{minipage}[b]{0.49\linewidth}
    \centering
    (a) w/o Prior
  \end{minipage}
  \begin{minipage}[b]{0.49\linewidth}
    \centering
    (b) w Prior
  \end{minipage}
  \caption{Effect of the uncertainty-aware objectness prior. False detections are effectively removed due to the objectness prior. With the Bayesian formulation (Eq.~\ref{eq:final_score}), the rescaled objectness prior yields more accurate detections.}
  \label{fig:prior-effect}
\end{figure}

In Eq.~\ref{eq:prob} and Eq.~\ref{eq:final_score}, $P_{\mathrm{scaled}}(O \mid p)$ acts as the prior, which represents the object likelihood in the proposal.
Then, $S_{\mathrm{joint}}(p,t)$ serves as the proxy likelihood, which approximates class-conditional evidence from the template similarity. %
Fig.~\ref{fig:prior-effect} demonstrates the effect of the objectness prior.
Specifically, with the objectness prior, many false positives shown in Fig.~\ref{fig:prior-effect}~(a) are removed, resulting in more accurate detections.

\begin{table*}[t]
\centering
\begin{tabular}{|c|l|ccccccc|c|c|}
\hline
\multicolumn{2}{|c|}{Method}  & LM-O & T-LESS & TUD-L & IC-BIN & ITODD & HB & YCB-V & $AP_{core}$ & Time \\ \hline \hline
\multirow{8}{*}{\rotatebox{90}{Detection}} 
& ZeroPose~\cite{chen2023zeropose}                 & 0.367 & 0.300 & 0.431 & 0.228 & 0.250 & 0.398 & 0.416 & 0.341 & 3.821 \\ 
& CNOS (SAM)~\cite{nguyen2023cnos}                 & 0.395 & 0.330 & 0.368 & 0.207 & 0.313 & 0.423 & 0.490 & 0.361 & 1.847 \\  
& CNOS (FastSAM)~\cite{nguyen2023cnos}             & 0.433 & 0.395 & 0.523 & 0.226 & 0.325 & 0.517 & 0.568 & 0.428 & \textbf{0.221} \\ 
& SAM6D(FastSAM)~\cite{lin2024sam}                 & 0.438 & 0.417 & 0.546 & 0.234 & 0.374 & 0.523 & 0.573 & 0.444 & 0.249 \\ 
& $\text{SAM6D}^{\dag}$(SAM)~\cite{lin2024sam}     & 0.465 & 0.437 & 0.537 & 0.261 & 0.394 & 0.530 & 0.518 & 0.449 & 2.795\\ 
& NIDS-Net~\cite{lu2024adapting}                   & 0.457 & 0.493 & 0.486 & 0.257 & 0.379 & 0.587 & 0.621 & 0.469 & 0.485 \\ 
&$\text{SAM6D}^{\dag}$(FastSAM)~\cite{lin2024sam}  & 0.463 & 0.458 & 0.573 & 0.245 & 0.419 & 0.551 & 0.589 & 0.471 & 0.445 \\ 
\cline{2-11} 
& MUSE (GD+SAM2)                                  & \textbf{0.511} & \textbf{0.494} & \textbf{0.601} & \textbf{0.337} & \textbf{0.490} & \textbf{0.616} & \textbf{0.684} & \textbf{0.533} & 0.505 \\ \hline
\multirow{9}{*}{\rotatebox{90}{Segmentation}}    
& ZeroPose~\cite{chen2023zeropose} & 0.356 & 0.337 & 0.421 & 0.293 & 0.210 & 0.453 & 0.534 & 0.372 & 3.821 \\ 
& CNOS (SAM)~\cite{nguyen2023cnos}  &  0.396 & 0.397 & 0.391 & 0.284 & 0.282 & 0.480 & 0.595 & 0.403 & 1.847 \\  
& CNOS (FastSAM)~\cite{nguyen2023cnos} & 0.397 & 0.374 & 0.480 & 0.270 & 0.254 & 0.511 & 0.599 & 0.412 & \textbf{0.221} \\ 
& SAM6D(FastSAM)~\cite{lin2024sam} & 0.406 & 0.393 & 0.501 & 0.278 & 0.290 & 0.522 & 0.606 & 0.428 & 0.249 \\ 
& $\text{SAM6D}^{\dag}$(FastSAM)~\cite{lin2024sam} & 0.422 & 0.420 & 0.517 & 0.293 & 0.319 & 0.548 & 0.621 & 0.449 & 0.445 \\ 
& SAM6D (SAM)~\cite{lin2024sam} & 0.444 & 0.408 & 0.498 & 0.346 & 0.300 & 0.557 & 0.595 & 0.450 & 	2.281 \\ 
& $\text{SAM6D}^{\dag}$(SAM)~\cite{lin2024sam}  & 0.460 & 0.451 & 0.569 & 0.357 & 0.332 & 0.593 & 0.605 & 0.481 & 	2.795 \\ 
    & NIDS-Net~\cite{lu2024adapting} & 0.439 & \textbf{0.496} & 0.556 & 0.328 & 0.315 & 0.620 & 0.650 & 0.486 & 0.485 \\ 
\cline{2-11} 
& MUSE (GD+SAM2)      & \textbf{0.477} & 0.478 & \textbf{0.573} & \textbf{0.433} & \textbf{0.391} & \textbf{0.635} & \textbf{0.690} & \textbf{0.525} & 0.505 \\ \hline
\end{tabular}
\caption{Comparison for BOP Classic-Core datasets. ${\dag}$ means using the RGB-D data. Our method has achieved the state-of-the-art performance in both detection and segmentation tasks. 'Time' is measured in seconds. Note that the boldface denotes best results.}
\vspace{-2mm}
\label{tab:classic_core_results}
\end{table*}

\section{Experiments}
\label{sec:experiments}

In this section, we present the evaluation of MUSE on the BOP Challenge 2025 benchmark across three official tracks: Classic-Core, H3, and Industrial. 
All experiments are conducted under a strict training-free setting, where target objects are never observed during training and no fine-tuning is applied. 
We begin by describing the experimental setup, evaluation metrics and dataset in implementation details. 
We then present a series of ablation studies to analyze the contribution of each component in MUSE, followed by quantitative and qualitative comparisons with recent state-of-the-art approaches.

\subsection{Implementation Details}
\label{subsec:implementation_details}

\noindent\textbf{Experimental setup}
We adapt Grounding DINO~\cite{liu2023grounding} with the Swin-B backbone, prompted by the generic text \texttt{"items"} to produce bounding boxes. 
These boxes are then passed as prompts to SAM2~\cite{ravi2024sam} with the Hiera-L backbone, which generates refined segmentation masks. 
Each candidate proposal is formed by combining the Grounding DINO bounding box with the SAM2 mask, ensuring that only the object region inside the box is preserved. 
This masked region serves as the final query proposal for subsequent matching.
Templates are generated from 3D CAD models rendered from 42 uniformly sampled viewpoints followed~\cite{nguyen2022templates} using vispy~\cite{campagnola2015vispy} library.
For both query and template feature extraction, we employ DINOv2~\cite{oquab2023dinov2} with the ViT-L variant as the extractor. 
By default, we set $e = 1.5$, $\alpha = 0.5$, $\beta = 0.8$, $\tau = 0.02$, and $\gamma = 0.1$. 
All experiments are performed on a single NVIDIA RTX 4090 GPU.

\vspace{3mm}
\noindent\textbf{Evaluation Metric}
We evaluate the method using Average Precision (AP), following the COCO and BOP challenge evaluation protocols~\cite{hodan2024bop, lin2014microsoft}. 
The AP is calculated by averaging the precision at Intersection over Union (IoU) thresholds ranging from 0.50 to 0.95 with an increment of 0.05.

\vspace{3mm}
\noindent\textbf{Dataset}
We evaluate our method on the seven BOP-Classic-Core datasets of the BOP benchmark~\cite{hodan2023bop}, including LM-O~\cite{brachmann2014learning}, T-LESS~\cite{hodan2017t}, TUD-L~\cite{hodan2018bop}, IC-BIN~\cite{doumanoglou2016recovering}, ITODD~\cite{drost2017introducing}, HomebrewedDB (HB)~\cite{kaskman2019homebreweddb}, and YCB-V~\cite{xiang2017posecnn}.  
We further evaluate on the three BOP-H3 datasets, HOT3D~\cite{banerjee2024introducing}, HOPEv2~\cite{tyree2022hope}, and HANDAL~\cite{guo2023handal}.  
Finally, we include the BOP-Industrial datasets, which consists of IPD~\cite{kalra2024towards}, XYZ-IBD~\cite{huang2025xyz}, and ITODD-MV~\cite{drost2017introducing}, to assess performance in challenging industrial scenarios.

\subsection{Ablation Study}
\label{sec:ablation}

We present a set of ablations to empirically validate the key components of our pipeline. 
Specifically, we investigate: (i) the effectiveness of feature integration with different pooling strategies,
(ii) the role of the joint similarity score, 
(iii) the impact of the uncertainty-aware objectness prior, 
and (iv) the additional design choices.

\vspace{3mm}
\noindent\textbf{Feature Integration and Pooling}
Table~\ref{tab:embedding-ablation} compares class-only, patch-only, and integrated embeddings under different pooling strategies. 
Class embeddings provide stable global alignment but lack local discriminativeness, yielding a baseline detection mAP of 0.478 with Grounding DINO and SAM2 Detector and Tanimoto Similarity. 
Patch embeddings capture fine-grained details, yet they are noisy and memory-intensive. As a result, performance appears to decline.
In contrast, our integrated approach with GeM pooling raises detection mAP to 0.496, consistently improving both $AP_{\text{core}}$ and $AP_{\text{industrial}}$ while requiring only a modest VRAM requirement. 
This demonstrates that GeM-based integration is critical for balancing global stability and local discriminativeness, with the additional advantage of memory efficiency.

\vspace{3mm}
\noindent\textbf{Joint Similarity Score.}
As described in Sec.~\ref{subsec:joint_similarity_score}, we combine absolute and relative similarities into a joint score. 
Without this step, detection performance plateaus at 0.496 mAP. 
Introducing the joint score further improves results to 0.505, as shown in Table~\ref{tab:ablation}. 
Figure~\ref{fig:joint-sim} illustrates that this score suppresses visually similar distractors and resolves ambiguities among confusing instances, translating directly into measurable accuracy gains.
This confirms that explicit relative reasoning across candidates is essential for robust detection.

\vspace{3mm}
\noindent\textbf{Uncertainty-Aware Objectness Prior}
Table~\ref{tab:ablation} shows that our Bayesian inference is effective, which we evaluate using the proposal's objectness confidence score.  
While the joint similarity alone achieves 0.505 mAP, incorporating the objectness prior boosts performance to 0.533, the highest score among all ablated variants. 
Figure~\ref{fig:prior-effect} shows how the prior reduces false detections from background regions. 
By rescaling the objectness distribution, we retain recall on true positives while suppressing false positives, leading to a principled and probabilistic final prediction. 
This demonstrates that proposal confidence carries indispensable prior information that must be incorporated in a Bayesian framework.

\vspace{3mm}
\noindent\textbf{Additional Design Choices}
Beyond feature integration and Bayesian inference, we also validate two additional design choices that contribute to the overall performance of MUSE. 
First, we replace the original proposal generation with a stronger detector by combining Grounding DINO with SAM2, inspired by the design of NIDS-Net~\cite{lu2024adapting}, which originally leveraged Grounding DINO with SAM. 
This substitution yields a substantial gain, boosting detection mAP from 0.361 to 0.469 and segmentation mAP from 0.403 to 0.474. 
Second, we replace the standard cosine similarity with Tanimoto similarity for feature matching, which better accounts for both vector direction and magnitude. 
This change further improves detection accuracy to 0.478 mAP, demonstrating the importance of adopting a more discriminative similarity metric.

\begin{table}[t]
\centering
\footnotesize
\begin{tabular}{l|c}
\toprule
Method & mAP (Detection / Segmentation) \\
\midrule
CNOS   & 0.361 / 0.403 \\
+ Grounding DINO + SAM2     & 0.469 {\tiny $\uparrow$ 0.108} / 0.474 {\tiny $\uparrow$ 0.069} \\
+ Tanimoto Similarity       & 0.478 {\tiny $\uparrow$ 0.009} / 0.483 {\tiny $\uparrow$ 0.009} \\
+ Feature Integration (GeM) & 0.496 {\tiny $\uparrow$ 0.018} / 0.498 {\tiny $\uparrow$ 0.015}\\
+ Joint Similarity Score    & 0.505 {\tiny $\uparrow$ 0.009} / 0.507 {\tiny $\uparrow$ 0.009} \\
+ Bayesian Prior (=MUSE)     & 0.533 {\tiny $\uparrow$ 0.028} / 0.525 {\tiny $\uparrow$ 0.018} \\
\bottomrule
\end{tabular}
\caption{Ablation study of MUSE. Results are reported as detection / segmentation mAP. Arrows indicate improvements relative to the previous setting.}
\label{tab:ablation}
\end{table}

\begin{figure*}[t]
    \centering

    \includegraphics[width=0.99\linewidth]{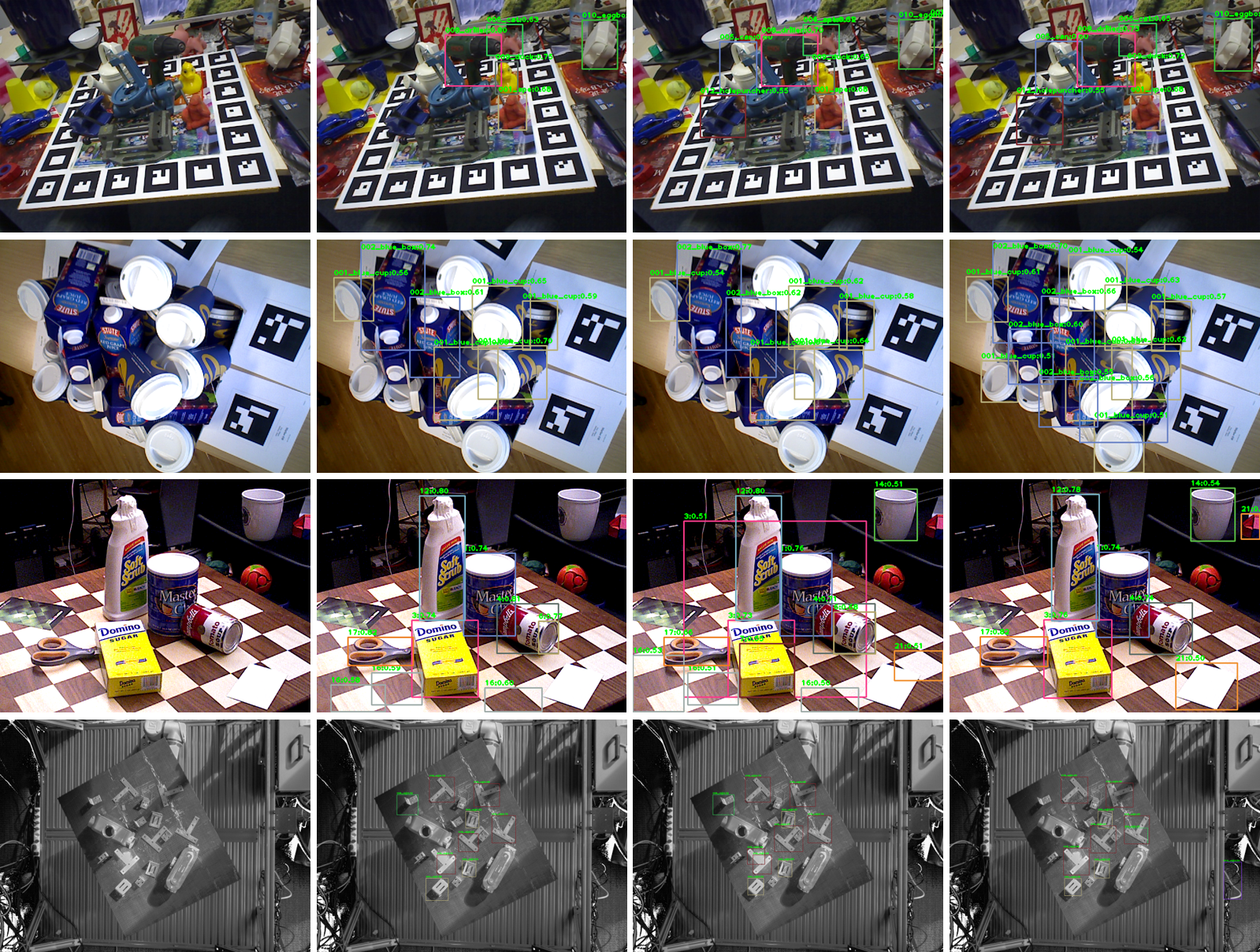}

    \vspace{0.3em}
    
    \begin{minipage}[b]{0.24\linewidth}
      \centering
      (a) Input Image
    \end{minipage}
    \begin{minipage}[b]{0.24\linewidth}
      \centering
      (b) CNOS (SAM)
    \end{minipage}
    \begin{minipage}[b]{0.25\linewidth}
      \centering
      (c) SAM6D (SAM)
    \end{minipage}
    \begin{minipage}[b]{0.24\linewidth}
      \centering
      (d) MUSE
    \end{minipage}
        
    \caption{Qualitative comparison of different algorithms on across various BOP-datasets. From top to bottom, LM-O, ICBIN, YCBV, IPD dataset. Each column shows results from (a) Input Image, (b) CNOS (SAM), (c) SAM6D (SAM), and (d) our proposed MUSE.} 
    \vspace{2mm}
    \label{fig:comparison}
\end{figure*}

\subsection{Comparison Results on BOP 2025}
\label{subsec:bop_results}

We compare our method MUSE with recent state-of-the-art model-based zero-shot approaches, 
including SAM6D~\cite{lin2024sam}, CNOS~\cite{nguyen2023cnos}, and NIDS-Net~\cite{lu2024adapting}, 
across all three official BOP 2025 Model-based tasks on unseen objects tracks: Classic-Core, H3, and Industrial. 

\vspace{3mm}
\noindent\textbf{BOP-Classic-Core}  
On the Classic-Core datasets (Table~\ref{tab:classic_core_results}), MUSE achieves the best overall performance in both detection and segmentation.  
For detection, MUSE reaches the highest $AP_{\text{core}}$ of 0.533, clearly surpassing the best baseline SAM6D (0.471).  
Similarly, in segmentation, MUSE attains the top $AP_{\text{core}}$ of 0.525, outperforming SAM6D (0.481) and NIDS-Net (0.486).  
Beyond accuracy, MUSE is also highly efficient, with an average runtime of only 0.505 seconds per image.  
While CNOS (FastSAM) runs faster (0.221s), MUSE is still the fastest among all remaining approaches.  
In particular, MUSE shows strong gains on challenging datasets like IC-BIN and ITODD, where densely packed and visually similar objects typically degrade performance.  
By leveraging our joint feature integration and Bayesian prior formulation, MUSE effectively disambiguates these cluttered scenarios, leading to robust improvements over prior methods.


\begin{table}[t]
\small
\resizebox{\columnwidth}{!}{%
\begin{tabular}{lccc|c|c c}
\toprule
Method & HOT3D & HOPEv2 & HANDAL & $AP_{\text{H3}}$ & Time (s) \\
\midrule
$\text{CNOS}$(SAM)      & 0.317 & 0.365 & 0.197  & 0.293  & 1.786 \\
$\text{CNOS}$(FastSAM)   & 0.350 & 0.313 & 0.246  & 0.303 & \textbf{0.332} \\
\midrule
$\text{MUSE}$(GD+SAM2)    & \textbf{0.438} & \textbf{0.460} & \textbf{0.357} & \textbf{0.418} & 0.919 \\
\bottomrule
\end{tabular}
}
\centering
\caption{Comparison of detection performance on BOP-H3 datasets. 
Results are reported as mean Average Precision (AP) on multiple subsets, with average inference time per image.}
\label{tab:h3_results}
\end{table}

\begin{table}[t]
\small
\resizebox{\columnwidth}{!}{%
\begin{tabular}{lccc|c|c c}
\toprule
Method  & IPD & XYZ-IBD & ITODD-MV & $AP_{\text{Industrial}}$ & Time (s) \\
\midrule
$\text{CNOS}$(SAM)                & 0.208 & 0.275 & 0.313 & 0.265 & 1.743 \\
$\text{SAM6D}^{\dag}$(FastSAM)    & 0.300 & 0.211 & 0.419 & 0.310 &\textbf{0.291}\\
$\text{SAM6D}^{\dag}$(SAM)        & \textbf{0.317} & 0.296 & 0.394 & 0.336 & 2.208 \\
\midrule
$\text{MUSE}$(GD+SAM2)            & 0.216 & \textbf{0.323} & \textbf{0.490} & \textbf{0.343} & 0.688 \\
\bottomrule
\end{tabular}
}
\centering
\caption{Comparison of detection performance on BOP-Industrial datasets. 
 ${\dag}$ means using the RGB-D data. 
Results are reported in terms of AP on multiple subsets, with average inference time per image.}
\label{tab:industrial_results}
\end{table}

\vspace{3mm}
\noindent\textbf{BOP-H3}  
In Table~\ref{tab:h3_results}, the H3 datasets, which includes HOT3D, HOPEv2, and HANDAL, 
MUSE establishes new state-of-the-art results with an overall $AP_{\text{H3}}$ of 0.418, 
surpassing CNOS (0.303) and CNOS (SAM) (0.293) by a large margin. 
In terms of efficiency, MUSE processes an image in 0.919 seconds on average. 
Although slower than CNOS (FastSAM) (0.332s), it is still considerably faster than CNOS (SAM) (1.786s). 
These results demonstrate that MUSE generalizes effectively across the diverse challenges of H3, 
covering egocentric hand–object interactions (HOT3D), cluttered household scenes (HOPEv2), 
and real-world manipulable tools (HANDAL).

\vspace{3mm}
\noindent\textbf{BOP-Industrial}  
In the Industrial datasets (Table~\ref{tab:industrial_results}), 
MUSE achieves the highest overall $AP_{\text{Industrial}}$ of 0.343, outperforming SAM6D$^{\dagger}$ (0.336), SAM6D$^{\dagger}$(FastSAM) (0.310), and CNOS (0.265).  
Despite relying only on RGB input, MUSE surpasses these RGB-D based approaches on challenging subsets such as XYZ-IBD (0.323 vs. 0.296 from SAM6D$^{\dagger}$) and ITODD-MV (0.490 vs. 0.419 from SAM6D$^{\dagger}$(FastSAM)), 
while IPD, with its structured-light depth modality, gives an advantage to RGB-D methods (0.216 vs. 0.317 from SAM6D$^{\dagger}$).
In terms of efficiency, MUSE also runs in 0.688 seconds per image, considerably faster than SAM6D$^{\dagger}$ (2.208s) while remaining competitive with SAM6D$^{\dagger}$(FastSAM) (0.291s). 
These results highlight that MUSE is effective for industrial scenarios, even only uses RGB data.


We provide a qualitative comparison in Fig.\ref{fig:comparison}.
Given the same input image (a), CNOS (b) detects multiple objects but often misses some objects or generates false positives.
SAM6D (c) improves detection coverage, yet still struggles with cluttered regions and produces noisy boundaries.
In contrast, our proposed MUSE (d) achieves the most consistent and accurate segmentation across diverse scenarios, successfully handling occlusion and complex backgrounds.
Overall, across Classic-Core, H3, and Industrial datasets, 
MUSE establishes the new state-of-the-art in model-based zero-shot detection and segmentation, demonstrating the effectiveness of our joint similarity score and uncertainty-aware prior.

\vspace{-2mm}

\section{Conclusion}
\label{sec:conclusion}

In this work, we suggest \textbf{MUSE} (\textbf{M}odel-based \textbf{U}ncertainty-aware \textbf{S}imilarity \textbf{E}stimation), a training-free framework for model-based zero-shot 2D object detection and segmentation.
We propose a new feature embedding scheme which integrates class and patch embeddings.
Specifically, the patch embeddings are normalized using the generalized mean pooling (GeM).
And we introduce a joint similarity score, which integrates an absolute score and a relative score.
Finally, we update the similarity score using an uncertainty-aware object prior. 
MUSE achieves state-of-the-art performance on the BOP Challenge 2025, ranking first in the Classic Core, H3, and Industrial tracks.
Therefore, we believe that MUSE is a promising framework for zero-shot 2D object detection and segmentation.

{
    \small

}

\end{document}